\documentclass[letterpaper, 10 pt, conference]{ieeeconf}  

\IEEEoverridecommandlockouts                              

\usepackage{cite}
\usepackage{amsmath} 
\usepackage{amssymb}  
\usepackage{xcolor}
\usepackage{algorithm}
\usepackage[noend]{algorithmic}
\usepackage{tikz}
\usetikzlibrary{bayesnet}
\usepackage{graphicx}
\usepackage{caption}
\usepackage{subcaption}
\usepackage{booktabs}   
\usepackage{multirow}

\makeatletter
\let\NAT@parse\undefined
\makeatother
\usepackage{hyperref}
\usepackage[capitalise]{cleveref}

\newcommand{\MM}{\mathcal{M}}

\newcommand{\XX}{\mathcal{X}}
\newcommand{\R}{\mathbb{R}}
\newcommand{\N}{\mathbb{N}}

\title{\LARGE \bf
  Learning Efficient Constraint Graph Sampling \\for Robotic Sequential Manipulation}

\author{Joaquim Ortiz-Haro$^{1, *}$, Valentin N. Hartmann$^{1, 3}$, Ozgur S. Oguz$^{1,2}$, Marc Toussaint$^{2,3}$
\thanks{The research has been supported by the German-Israeli Foundation for Scientific Research (GIF) grant I-1491-407.6/2019, the International Max-Planck Research School for Intelligent Systems (IMPRS-IS), and the German Research Foundation (DFG) under Germany's Excellence Strategy -- EXC 2120/1 -- 390831618.}
\thanks{$^{*}${\tt\small joaquim.ortiz@ipvs.uni-stuttgart.de}}%
\thanks{$^{1}$Machine Learning \& Robotics Lab, University of Stuttgart, Germany}%
\thanks{$^{2}$Max Planck Institute for Intelligent Systems, Germany}%
\thanks{$^{3}$Learning and Intelligent Systems Lab, TU Berlin, Germany}
}

\begin{document}

\maketitle
\thispagestyle{empty}
\pagestyle{empty}

\begin{abstract}

Efficient sampling from constraint manifolds, and thereby generating a diverse set of solutions for feasibility problems, is a fundamental challenge.
We consider the case where a problem is factored, that is, the underlying nonlinear program is decomposed into differentiable equality and inequality constraints, each of which depends only on some variables.
Such problems are at the core of efficient and robust sequential robot manipulation planning.
Naive sequential conditional sampling of individual variables, as well as fully joint sampling of all variables at once (e.g., leveraging optimization methods), can be highly inefficient and non-robust.
We propose a novel framework to learn how to break the overall problem into smaller sequential sampling problems.
Specifically, we leverage Monte-Carlo Tree Search to learn assignment orders for the variable-subsets, in order to minimize the computation time to generate feasible full samples.
This strategy allows us to efficiently compute a set of diverse valid robot configurations for mode-switches within sequential manipulation tasks, which are waypoints for subsequent trajectory optimization or sampling-based motion planning algorithms.
We show that the learning method quickly converges to the best sampling strategy for a given problem, and outperforms user-defined orderings or fully joint optimization, while providing a higher sample diversity.
\urlstyle{same}
Video: 
\url{https://youtu.be/mCNdvjTbHNI}

\end{abstract}

\section{Introduction}

Highly constrained problems are at the core of a wide variety of different applications  (e.g. scheduling \cite{johnston1994spike}, hardware verification \cite{dechter2002generating}, or robotic manipulation planning \cite{toussaint2018differentiable}). 
To robustly tackle such problems, they are represented as constrained optimization or satisfaction problems, and one aims to find a diverse set of feasible solutions.

Focusing on problems in continuous domains, there are two main approaches to generate solutions: 
The first is to use a nonlinear solver to generate a joint assignment for all variables simultaneously.
If the problem is highly nonlinear, however, joint optimization is sensitive to the initial guess, and is prone to converging to an infeasible point, or repeatedly to the same local optima.
The second direction is to decompose the problem using sequential assignment of subsets of variables. 
However, this approach is often impossible due to joint constraints on variables.
Defining dedicated constraint manifold sampling operations and a sampling order to address this issue requires in-depth knowledge of the problem domain.
While defining good decompositions is feasible for simple or structured settings, in general, they are suboptimal for complex scenarios.

\begin{figure}
\centering
\begin{subfigure}[t]{.23\textwidth}
    \centering
    \includegraphics[width=.97\linewidth]{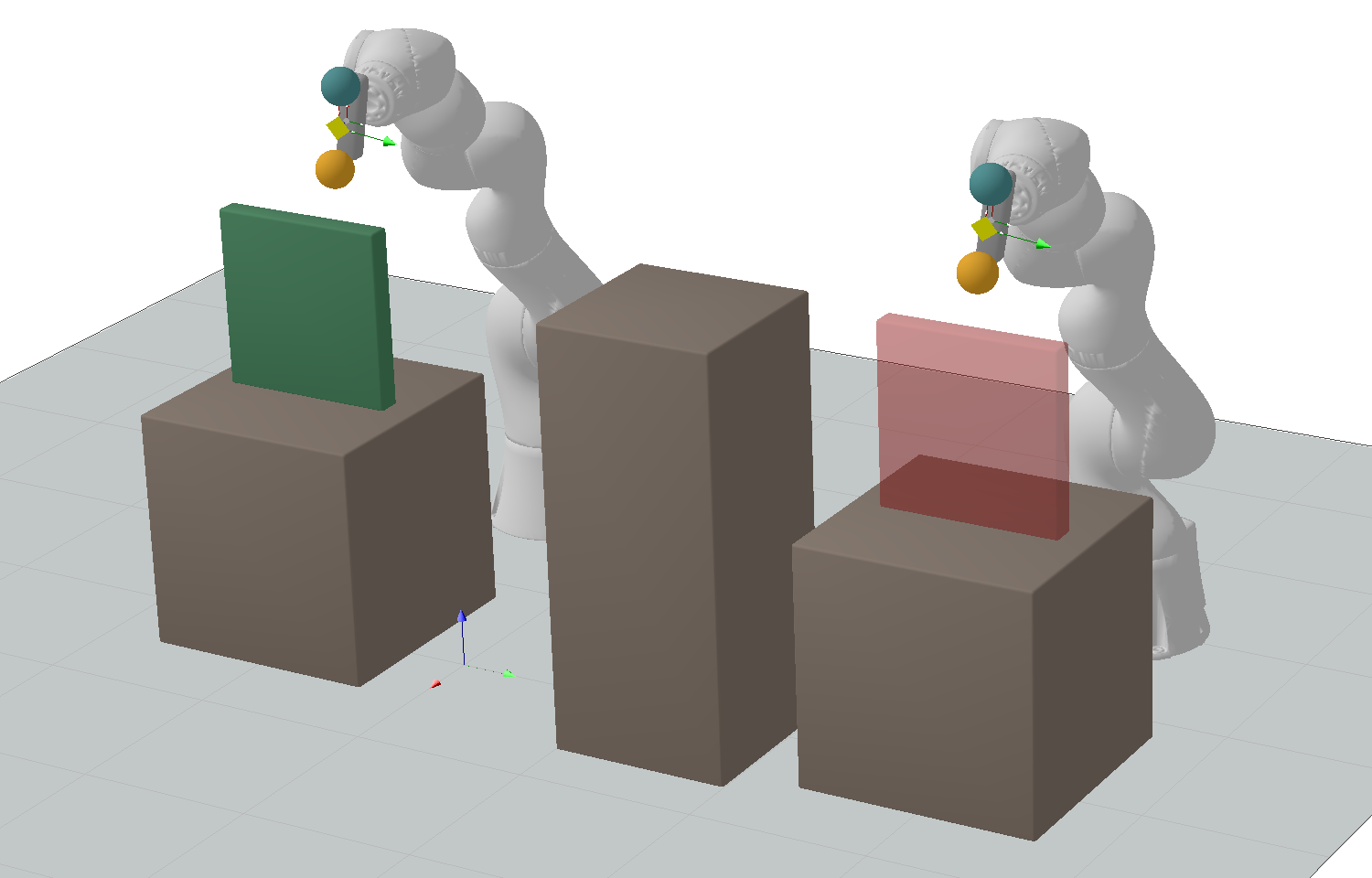} 
\end{subfigure}%
\begin{subfigure}[t]{.23\textwidth}
    \centering
    \includegraphics[width=.97\linewidth]{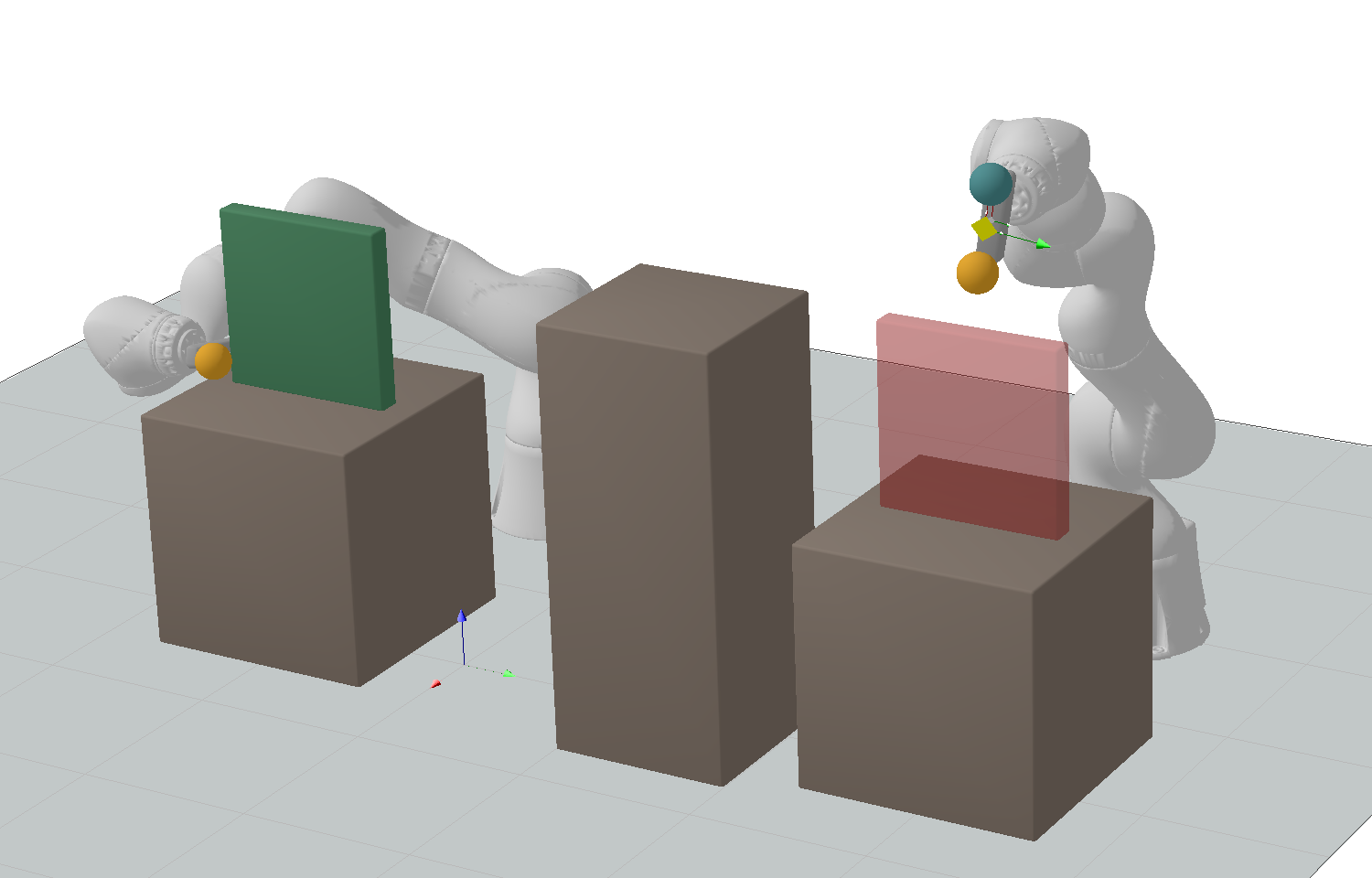} 
\end{subfigure}\\[1mm]

\begin{subfigure}[t]{.23\textwidth}
    \centering
    \includegraphics[width=.97\linewidth]{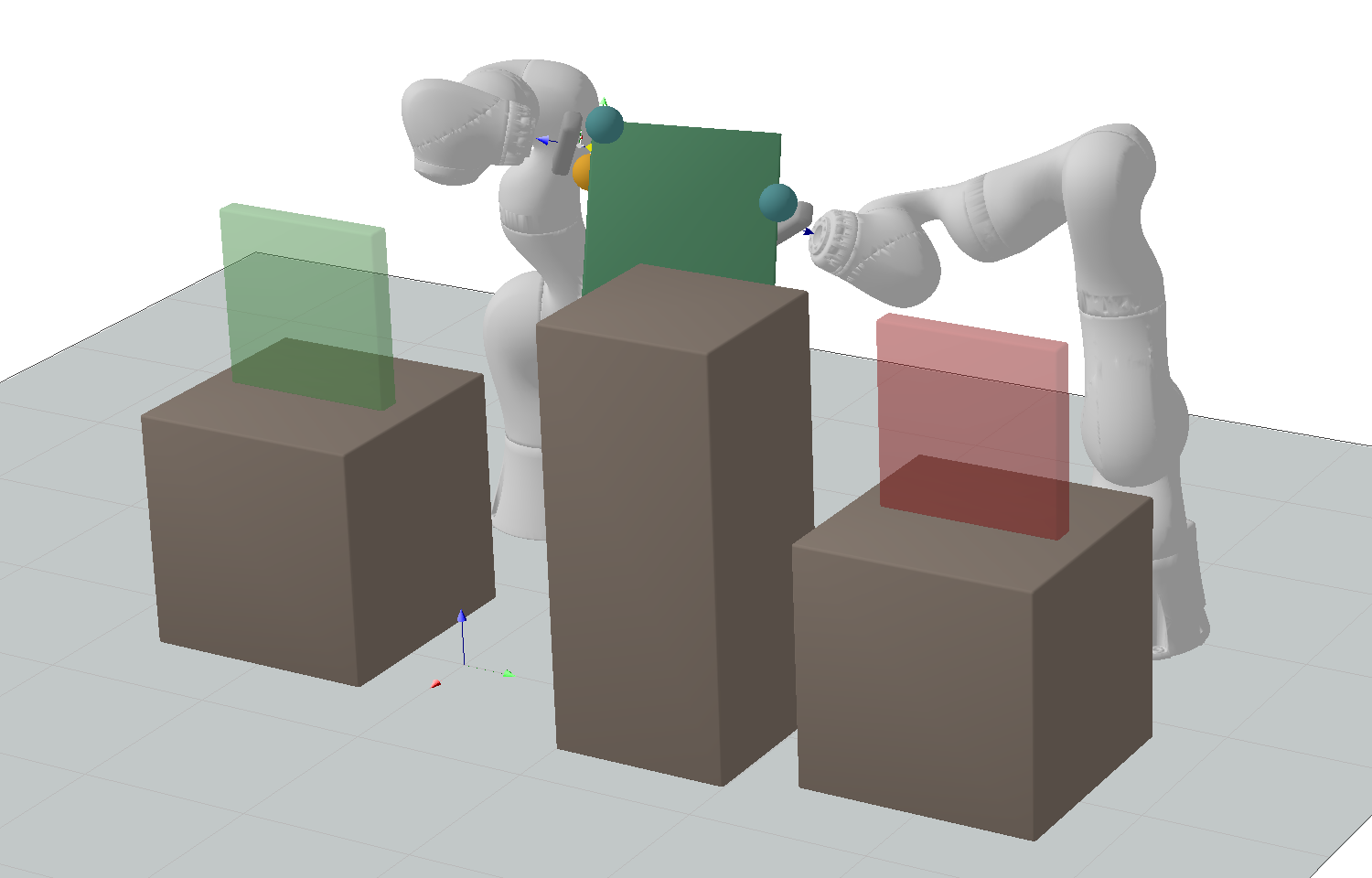} 
\end{subfigure}%
\begin{subfigure}[t]{.23\textwidth}
    \centering
    \includegraphics[width=.97\linewidth]{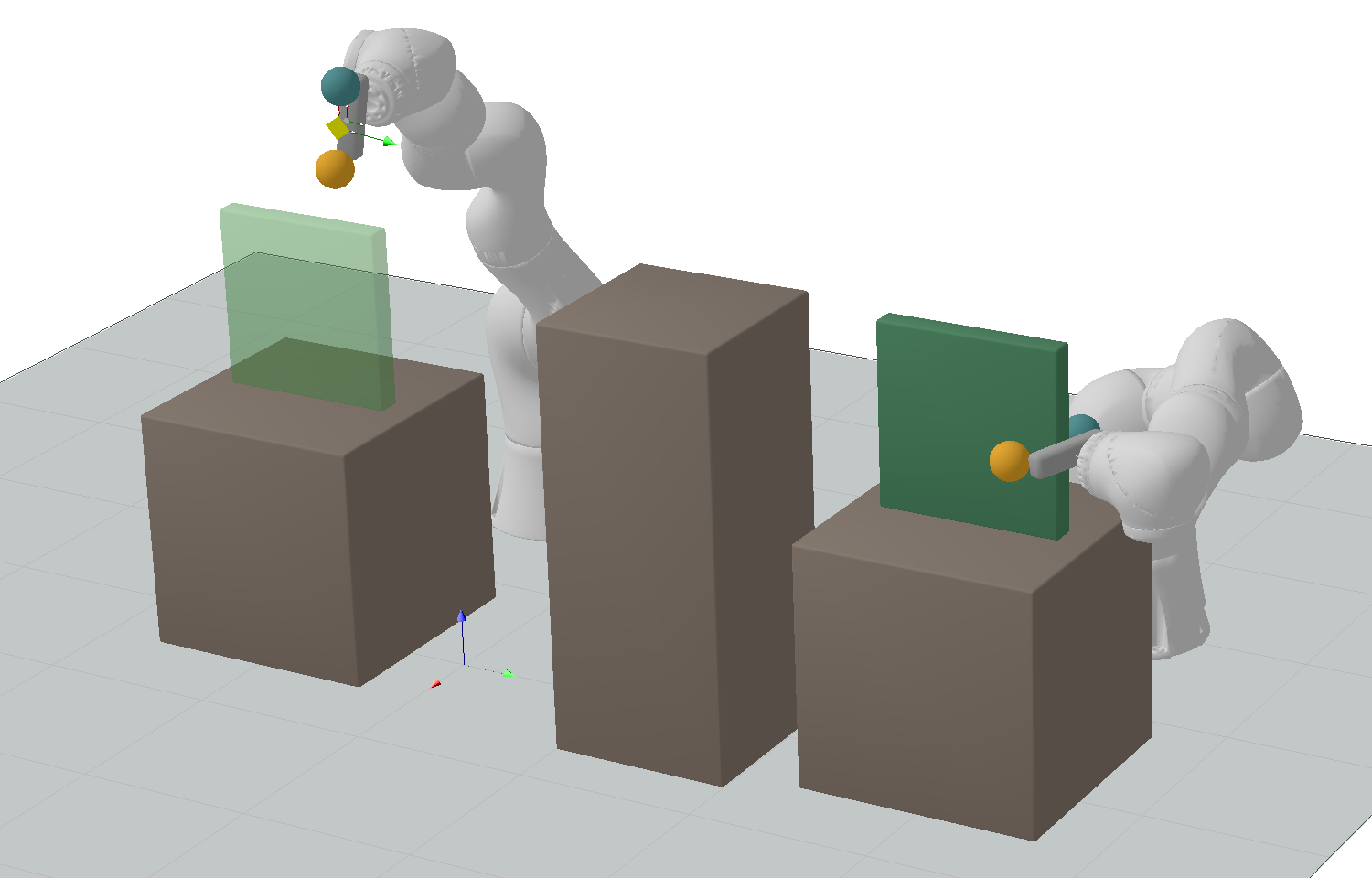} 
\end{subfigure}
  \caption{A sequence of mode-switches in the \textit{Handover} problem. Goal: Move the green box to the target location (red). Robot A picks the box (top-right) and hands it to robot B (bottom-left) to place it in the goal position (bottom-right).}
    \label{fig:keyframes-handover}
\end{figure}

We demonstrate our work on problems arising in robotic sequential manipulation, which requires finding feasible solutions that satisfy the kinematic and dynamic constraints imposed by the robot and the environment, possibly arising from future actions and goals. 
Such problems implicitly encode discrete structures due to the stable interaction modes that describe the contact- or kinematic switches.  
These mode-switches are discrete snapshots of the configuration variables when the constraint activity changes, e.g., when the robot picks up an object (\cref{fig:keyframes-handover}).
Generating a diverse set of feasible mode-switches, and then optimizing the  trajectories while considering the whole set
can alleviate infeasibility and suboptimality issues that occur when only using one possible mode-switching sequence.

As example, consider the mode-switch configurations of a pick-and-place problem, defined with three sets of variables: the robot configurations (joint values) when
picking and 
placing the object, and 
the relative transformation between the gripper and the object. 
Some possible sampling sequences to generate a full sample are:
(i) optimize all variables jointly, (ii) sample the relative transformation first, and then compute the robot configurations, or (iii) compute the pick configuration and relative transformation first, and then the place configuration.




In this work, we present an algorithm to efficiently generate a diverse set of solutions for high-dimensional nonlinear programs containing many infeasible local optima by exploiting its factored structure. 
Our method relies on the factored graph representation of the task to reason online about the best sequence of constrained conditional sampling operations. 




Thus, our approach learns a decomposition based on the given microscopic factorization, optimizing for the success rate and computation time of sampling.
This maximizes the number of solutions that fulfill the constraints in a fixed computational time.



Our main contributions are:
\begin{itemize}
    \item A new framework to model constrained conditional sampling in constraint graphs and computation operations as a stochastic decision process,
    \item Pruning-strategies that leverage the factored constraint graph representation of the nonlinear program to reduce the space of possible computations, and
    \item An algorithm based on Monte-Carlo Tree Search to find sequences of computational operations that maximize the number of generated solutions in a fixed computational time.
\end{itemize}




\section{Related Work}
\subsection{Problem Decomposition and Constraint Graphs}
In continuous optimization, decompositions using the underlying problem structure are used to improve the performance of the algorithms both from a theoretical and practical perspective~\cite{dantzig1960decomposition,bnnobrs1962partitioning, boyd2011distributed,bertsekas1979convexification}. 
Examples in robotic applications demonstrate attempts to decompose complex nonlinear problems in a sequence of easier subproblems, where the solution of the simpler problems is used to guide optimization and sampling methods towards the solution of the full problem~\cite{19-driess-RSSws, 17-toussaint-ICRA, tonneau2018efficient, orthey2018quotient}.


In constraint satisfaction problems (CSP), constraint graphs are often used to reason about the problem structure, and to decompose problems into simpler ones.
Algorithms solving CSPs make use of the graph structure, e.g. identifying connected components or trees, to efficiently assign variables \cite{mouhoub2011heuristic}. 
A general overview on CSPs can be found in \cite{rossi2006handbook}.
There is a rising interest for sampling solutions satisfying CSPs \cite{dechter2002generating, gogate2006new, ermon2012uniform}.
Generating multiple solutions to the problem at hand improves the applicability of these methods in settings where some constraints can not be modeled or only evaluated a posteriori \cite{danna2007generating}. 


However, to the best of our knowledge, the current literature related to the problem of sampling from a CSP focuses mostly on CSPs with finite, discrete domains.


In this work, we use the problem structure of the continuous constraint satisfaction problem in the form of a constraint graph to reduce the set of possible computation sequences.
We reason about the best way to generate \textit{multiple diverse} samples,
instead of only generating a \textit{single solution} to a constraint satisfaction problem.





\subsection{Meta-Decision Processes}

The optimization over computational operations relates directly to optimal meta-decision-making  \cite{russell1991principles}. 
Meta-decision-making has various applications in heuristic search~\cite{o2015metareasoning, lieder2014algorithm, zilberstein2011metareasoning, karpas2018rational}, and other resource bound planning processes~\cite{bratman1988plans, boddy1989solving}.
More recently, meta-reasoning has been applied in reinforcement learning~\cite{pascanu2017learning}, or in temporal planning~\cite{cashmore2018temporal} amongst others.
The robotics community is applying similar ideas in their domain, namely in path planning.
In~\cite{mandalika2019generalized}, deciding which edge to evaluate next in a lazy motion planner is explicitly regarded as decision problem.

We use bandit algorithms~\cite{auer2002finite,kocsis2006bandit} to reason about the sequence of computational operations to generate solutions to a nonlinear program.
The decisions we take are the discrete choices of which variables should be computed and assigned next.



\subsection{Robotic Sequential Manipulation}

Since we demonstrate our contributions on sequential manipulation problems, we briefly give an overview over the approaches taken to solve such problems.
Task and motion planning (TAMP) is concerned with finding action sequences, and corresponding motion plans to fulfil some symbolic and/or geometric goal~\cite{sucan-kavraki2011on-advantages-of,dantam2016tmp,lagriffoul2014efficiently}.

A comprehensive review on the history and current developments of TAMP can be found in~\cite{garrett2020integrated}.
In the following, we highlight some work that is specifically relevant for us: 
Logic geometric programming~\cite{toussaint2018differentiable,17-toussaint-ICRA,driess2020deep, driess2021DVRC} solves several optimization problems, of which one tries to find the mode switches for a previously determined action sequence.
In~\cite{garrett2018sampling}, the full TAMP problem is regarded as a factored transition graph, which is then sampled to find solutions to the whole problem.
Multi-modal motion planning~\cite{hauser2010multi, Kingston2020a} extends motion planning to solve manipulation planning~\cite{berenson2011task, kingston2018sampling} by incorporating mode-switches as the intersection of overlapping manifolds.  




 Generally, current TAMP approaches assume that mode-switches  can be efficiently generated with either joint optimization or user defined sampling sequences.  
In our work, we focus on problems where this assumption does not hold due to obstacles, nonlinear constraints, and strong dependencies between variables, and present a method to efficiently sample the solution manifold for such cases.

Our method can be directly integrated as an inner module in TAMP approaches, enabling them to solve more challenging and constrained problems. In fact, in cluttered scenarios such as a construction site, generating the mode-switches becomes one of the computational bottlenecks of the whole TAMP pipeline~\cite{20-hartmann-IROS}. 




\section{Constraint Graph Sampling}
\begin{figure*}[t]
\centering

\begin{subfigure}[b]{.2\textwidth}
  \centering
\begin{tikzpicture}[scale=0.7,every node/.style={transform shape}]
        \node[latent] (t) {$t$} ; %
        \node[latent, below=of t, xshift=-.5cm] (q1) {$q_1$} ; %
        \node[latent, below=of t, xshift=.5cm] (q2) {$q_2$} ; %
        \factor[above=of q2] {Kina2} {left:Kin} {t,q2} {};
        \factor[above=of q1] {Kina1} {left:Kin} {t,q1} {};
        \factor[below=of q1] {Collision} {left:Collision} {q1} {};
        \factor[below=of q2] {Collision} {right:Collision} {q2} {};
        \factor[above=of t] {Grasp} {Grasp} {t} {};
  \end{tikzpicture}
  \caption{Pick and Place}
  \label{fig:pick_place}
\end{subfigure}%
\hspace{1em}
\begin{subfigure}[b]{.38\textwidth}
  \centering
    \begin{tikzpicture}[scale=0.7,every node/.style={transform shape}]
        \node[latent ] (p) {$p$} ; %
        \node[latent, left=1.5 of p , yshift=.0cm] (ta) {$t_a$} ; %
        \node[latent, right=1.5 of p, yshift=.0cm ] (tb) {$t_b$} ; %
        \node[latent, below=of ta, xshift=-.5cm] (qa1) {$q_{a_1}$} ; %
        \node[latent, below=of ta, xshift=.5cm] (qa2) {$q_{a_2}$} ; %
        \node[latent, below=of tb, xshift=-.5cm] (qb2) {$q_{b_2}$} ; %
        \node[latent, below=of tb, xshift=.5cm] (qb1) {$q_{b_1}$} ; %


        \factor[above=of qa2] {Kina2} {left:Kin} {ta,qa2,p} {};
        \factor[above=of qb2] {Kinb2} {left:Kin} {tb,qb2,p} {};

        \factor[above=of qa1] {Kina1} {left:Kin} {ta,qa1} {};
        \factor[above=of qb1] {Kinb1} {left:Kin} {tb,qb1} {};

        \factor[below=1.5 of p] {Collision} {below:Collision} {qa2,qb2} {};
        \factor[below=of qa1] {Collision} {left:Collision} {qa1} {};
        \factor[below=of qa2] {Collision} {right:Collision} {qa2} {};
        \factor[below=of qb1] {Collision} {right:Collision} {qb1} {};
        \factor[below=of qb2] {Collision} {left:Collision} {qb2} {};
        \factor[above=of p] {Position} {left:Position} {p} {};
        \factor[above=of ta] {Grasp} {Grasp} {ta} {};
        \factor[above=of tb] {Grasp} {Grasp} {tb} {};
  \end{tikzpicture}
  \caption{
    Handover
}
  \label{fig:factor-handover}
\end{subfigure}%
\hspace{1em}
\begin{subfigure}[b]{.38\textwidth}
  \centering
\begin{tikzpicture}[scale=0.7,every node/.style={transform shape}]
        \node[latent ] (p) {$p$} ; %
        \node[latent, left=2 of p , yshift=.0cm] (ta) {$t_a$} ; %
        \node[latent, right=2 of p, yshift=.0cm ] (tb) {$t_b$} ; %
        \node[latent, below=of ta, xshift=-.5cm] (qa1) {$q_{a_1}$} ; %
        \node[latent, below=of ta, xshift=.5cm] (qa2) {$q_{a_2}$} ; %

        \node[latent, below=of p, xshift=-.5cm] (qx) {$q_x$} ; %
        \node[latent, below=of tb, xshift=-.5cm] (qb1) {$q_{b_1}$} ; %
        \node[latent, below=of tb, xshift=.5cm] (qb2) {$q_{b_2}$} ; %
        \node[latent, below=of p, xshift=.5cm] (ta2) {$t_{x}$} ; %


        \factor[above=of qa2] {Kina2} {left:Kin} {ta,qa2,p} {};
        \factor[above=of qb1] {Kinb2} {left:Kin} {tb,qb1,p,ta2} {};

        \factor[above=of qa1] {Kina1} {left:Kin} {ta,qa1} {};
        \factor[above=of qb2] {Kinb1} {left:Kin} {tb,qb2,ta2,p} {};

        \factor[below=of p] {KinX} {left:Kin} {ta2,qx,p} {};

        \factor[below=of qa1] {Collision} {left:Collision} {qa1} {};
        \factor[below=of qa2] {Collision} {right:Collision} {qa2} {};
        \factor[below=of qb2] {Collision} {right:Collision} {qb2} {};
        \factor[below=of qb1] {Collision} {left:Collision} {qb1} {};
        \factor[below=of qx] {Collision} {right:Collision} {qx} {};
        \factor[above=of p] {Position} {left:Position} {p} {};
        \factor[above=of ta] {Grasp} {Grasp} {ta} {};
        \factor[right=of ta2] {Grasp} {below:Grasp} {ta2} {};
        \factor[above=of tb] {Grasp} {Grasp} {tb} {};
  \end{tikzpicture}
  \caption{Banana}
  \label{fig:banana-with-regrasp}
\end{subfigure}%
\caption{
Constraint graphs in manipulation planning, \textit{Pick-and-place} is used to illustrate how the space of possible computations can be efficiently pruned. We also used two complex manipulation scenarios, \textit{handover} and \textit{banana}, to evaluate our method.}
\label{fig:factor_graphs}
\end{figure*}
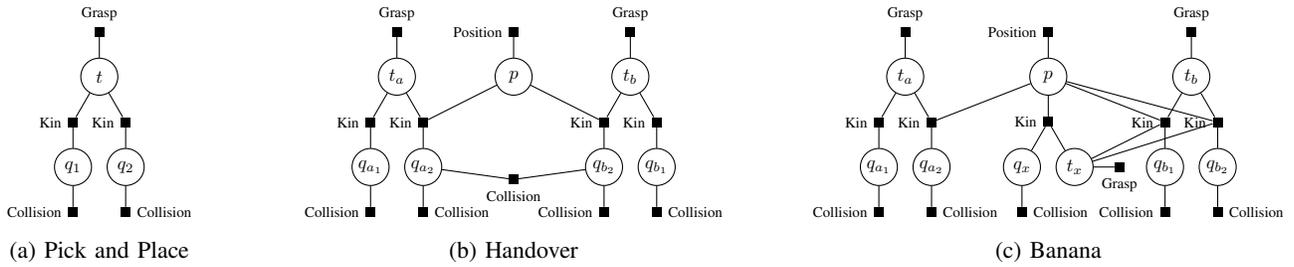

\subsection{Constraint Sampling Program Formulation}
Consider a nonlinear program (NLP) without cost term in its general form, 
\begin{equation}\label{eq:general_problem}
  \text{find} \  x \in \mathcal{X} \  \text{s.t.}  \ h_{\textrm{ineq}}(x) \le 0,\ h_{\textrm{eq}}(x) = 0 
\end{equation}
where $x$ are variables in some continuous space $ \XX $ of dimensionality $k$ (e.g., the vector space $ \R^k$), $h_{\textrm{eq}} : \mathcal{X} \to \R^{m_{\textrm{eq}}}$ and $h_{\textrm{ineq}} : \mathcal{X} \to \R^{m_{\textrm{ineq}}}$ are differentiable vector functions representing the constraints.

Let $\MM  =  \{ x \in \mathcal{X} : h_{\textrm{ineq}}(x) \le  0, h_{\textrm{eq}}(x)=0 \}$ be the feasible set of \eqref{eq:general_problem}: a nonlinear and possibly disconnected  manifold. Our goal is to generate a diverse set of  discrete samples $ \{ x^{(i)} : x^{(i)} \in \MM \} $ that is an $\epsilon$-cover. 
An $\epsilon$-cover is a finite set of samples  $ \{ x^{(i)} : x^{(i)} \in \MM \} $ s.t. $\max_{x \in \MM} \min_{x^{(i)}} \| x - x^{(i)} \|  < \epsilon$. 



In general, a solution $x^{(i)}$ can not be generated with a closed-form sampling expression.
Thus, the standard approach to generate a solution is to sample on the ambient space $x_s \in \mathcal{X} $  and project it to $\MM$ with a local optimizer. 


\subsection{Constraint Graphs}
We use the knowledge about the problem structure (variables, constraints and their dependencies) to transform the general formulation of the NLP into a constraint graph:
We assume the space $\XX = \XX_1 \times \ldots \times \XX_n$ (and thus $x = \{x_1, ..., x_n\}$) can be factored into $n$ variables.

Similarly, the constraints $h_{\textrm{eq}}(x), h_{\textrm{ineq}}(x)$ are factored into $l$ sets of constraints, $\{ h_1, h_2 , \ldots , h_l \} \quad \text{with} \quad h_j : \R^{a_j} \to \R^{b_j}$ where $a_j,\ b_j \in \N$ are the dimensions of the domain and co-domain of the constraint $h_j$, and each constraint $h_j$ depends only on a subset of variables.
Such factorizations naturally arise in many applications, where each variable has some semantic and geometric meaning. 

Using this factorization, the problem can be represented graphically as a constraint graph~\cite{rossi2006handbook}, which are a special instance of factor graphs~\cite{frey1997factor}.

We show examples of such constraint graphs for robotics manipulation in \cref{fig:factor_graphs}.
The problems involve robot configurations $q$, object positions $p$ and relative transformations $t$ between the robotic gripper and the object.
These variables are constrained by the kinematic chain of the robot (\textit{Kin}), collision avoidance (\textit{Collision}), grasping (\textit{Grasp}), and position constraints (\textit{Position}).
By combining these basic blocks, we can model complex tasks such as handovers or sequential manipulation with intricate dependencies.

Our approach can readily also handle cost objectives in the original NLP (\ref{eq:general_problem}), at which point we would more naturally speak of factor graphs.
Note that, while cost objectives can be included, from here on, we focus on feasibility problems.

A key observation is that constraint graphs are typically sparse, with few edges (i.e., each constraint is only applied to a limited number of variables). 
This property will be exploited in the next section to design efficient algorithms.

\section{Sequential Sampling in Constraint Graphs as a Markov Decision Process}
\label{sec:framework}

We first observe that a complete assignment of the variables $x=\{x_1,\ldots,x_n\}$ can be computed in different ways: Variables could be assigned jointly, one by one, or following a sequence of blocks of different sizes.
We define the computational state $s\in P(S)$ to refer to the indices of $x$ that have been assigned, thus $S$ denotes the set of indices $\{1, ..., n\}$, with $P$ being the powerset-operator.
Hence, any full assignment for $x$ can be computed following a sequence of assignments $H = \left(s_0,  s_1, \ldots, s_g\right)$ where $s_i \subset s_{i+1}$, and $s_0=\{ \}, s_g = S$. 
We use $H_i=(s_0, ..., s_i)$ to denote a partial sequence up until the $i$-th assignment.

A transition $s_i \to s_j$ in a sequence implies a computational operation $x_{s_j} = o_{s_i, s_j}(x_{s_i})$ that assigns numerical values to the new variables, given $x_{s_i}$, which is kept fixed.
We call a transition \textit{valid}, if $s_i \subset s_j$.
Thus, $\mathcal{Y}(s_i) = \{s_{j}: s_i\subset s_{j}\}$ denotes the states that can be reached with a valid transition from $s_i$.
Generally, the transition operations are implemented either with direct conditional sampling or using optimization methods initialized with a randomized guess.  \\
For instance, in the pick-and-place problem presented in the introduction, examples of such transitions are generating 6-DOF grasps or solving inverse kinematics (i.e.  computing robot joint values for a given end-effector position). 



\Cref{alg:generate-solutions} shows how to generate solutions from a constraint graph by choosing a valid transition and associated operation at each step, and therefore fixing the sequence of assignments.
The performance of the algorithm depends heavily on how transitions are selected (line 8 -- our method described in \cref{sec:UCT} is in orange).

Each transition $s_i \to s_j$ is a conditional sampling operation, which generates samples with an unknown but defined conditional probability density 
$p_{s_j | s_i}(x_{s_j} | x_{s_i})$, defined on the feasible manifold of $x_{s_j}$.
There are no assumptions on the shape of the distribution. 
We assume that the probability density is zero everywhere in case the transition conditioned on $x_{s_i}$ is not feasible. 
The joint probability density $p_{H_j}(x_{s_j}) = p_{s_j | s_i}(x_{s_j}  | x_{s_i})  p_{H_i}(x_{s_i})$ thus depends on the history $H_j=(s_0,..,s_i,s_j)$ of transitions to reach $s_j$.



A valid transition $s_i\to s_j$ between computational states does not always produce an assignment that satisfies the associated constraints for $x_{s_j}$.
There are two sources of infeasibility: \textit{(i)} the problem conditioned on the previously assigned variable $x_{s_i}$ could be infeasible, or \textit{(ii)} the sampling operation could fail to generate a solution even if one exists.

We can define an ad hoc success rate estimate for transitions between the current sequence $H_i$ and the next computational state $s_{i+1}$ (equivalently, between $H_i$ and $H_{i+1} = H_{i} \cup s_{i+1} $) as follows:
\begin{align}\label{eq:successRate}
  Pr \left( H_i  \to H_{i+1} \right) = 
  \hat p_{H_i, H_{i+1}} \ \frac{ \int_{ \Omega_{i,i+1 } } p_{H_i}(x_{s_{i}}) d\Omega }{ \int_{ \Omega_i }  p_{H_i}(x_{s_i}) d\Omega },
\end{align}
where $\Omega_i$ is the feasible space for  $x_{s_i}$ and  $\Omega_{i,i+1} \subseteq \Omega_{i} $ is the feasible space for $x_{s_i}$ for which a joint sample $x_{s_{i+1}}$ exists.
In this estimate, the success rate $\hat p_{H_i, H_{i+1}} \in (0,1)$ represents the probability that a solution satisfying the constraints will be found if one exists, which we estimate as constant for a given transition, rather than depending on the previous assignment $x_{s_i}$.
Note that the success rate could be low for certain transitions, while it could be close to 1 for others.





\subsection{Computational State Transition as a Markov Decision Process}
\label{sec:markov}


We can now define a Markov Decision Process (MDP):
The state space is $\{ H_i \} \cup \{\neg\}$, i.e., the sets of partial sequences starting from $s_0$ plus an infeasible state $\neg$.
The action space contains the possible transitions between sequences, i.e., adding a new computational state.
Each action has a success rate $p_a = Pr(H \to H')$ of reaching the new sequence as defined in (\ref{eq:successRate}), (i.e.\ the probability of generating an assignment satisfying the constraints) and probability $1-p_a$ of going to the infeasible state $\neg$.
We note that the MDP has a maximum horizon length of $n$ transitions, i.e., the size of the variable set, which is equivalent to assigning all the variables one by one.

We use the reward structure $R$ to obtain a concept of optimality that directly relates to our objective of maximizing the number of samples:
We introduce a reward $r_g=1$ for reaching the goal, i.e., reaching $S$, and a stochastic cost, i.e., a negative reward, $r_t$ on each transition, which is the time the transition takes.
The cost $r_t$ is not only added at the final node of the sequence, but directly at the intermediate nodes.
We weigh the two costs linearly, obtaining $\lambda r_t$ and $(1-\lambda) r_g$ with $\lambda \in (0,1)$.





\section{Choosing Computational operations with Monte-Carlo Tree Search}
We use Monte-Carlo Tree Search (MCTS, \cite{browne2012survey}) on the previously defined MDP to find the optimal sequence of computational operations. 
MCTS incrementally builds a tree of possible transition sequences to find the most promising.
This is achieved by randomly choosing transitions starting from the initial state $s_0$ of the MDP, 
executing its corresponding conditional sampling operation 
and weighting the generated sequences using the obtained reward.
Thus, in our setting, the nodes of the tree built by the MCTS are sequences of computational states.

The reward structure of the MDP guides the tree search algorithm to choose transitions that have a low computational cost and high probability of producing a full sample.
In practice, under the assumption that $\lambda$ is chosen well\footnote{An estimate of the average computational cost $\hat{c} =- \sum {r_t}$ can be used as a guidance for choosing a suitable $\lambda = \frac{1}{\hat{c}+1}$.}, this means that the number of generated samples is maximized.

\subsection{Upper Confidence Tree (UCT)}\label{sec:UCT}
In particular, we use the UCT algorithm \cite{kocsis2006bandit} to balance the exploitation of known sequences with the exploration of new sequences.
UCT expands the child node $i$ that maximizes
\begin{equation}\label{eq:uct}
    Q_i + c\sqrt{\frac{\ln N_i}{n_i}}, 
\end{equation}
where $Q_i$ is the current estimate of the expected reward of node $i$, $N_i$ is the number of simulations that evaluated the parent node, $n_i$ is the number of rollouts that evaluated node $i$, and $c$ is a constant chosen by the user.
The estimate of the $Q$ function of the nodes of the trees converges as the number of rollouts increases. 
Hence, the UCT algorithm incrementally builds a search tree, estimates the $Q$ values of each transition, and chooses actions using the upper confidence bound~\eqref{eq:uct}, leading to \cref{alg:generate-solutions}.

We note that, if at least one of the sampling sequences assigns non zero probability density to the whole feasible manifold, so does our algorithm, since UCT never stops exploring all possible sequences.

\begin{algorithm}[t] 
\caption{Framework to generate solutions from a constraint graph, with our contributions in {\color{orange}orange}.}
\label{alg:generate-solutions} 
\begin{algorithmic}[1] 
   \STATE \textbf{Input:} Constraint graph,  time budget $T$. 
   \STATE { $L = \{ \}$  } (empty list of samples)
    \WHILE{$\text{current time} \  t \le T$}
    \STATE{feasible = True, $x = \{ \}$  (empty sample)}
    \STATE{ $s = s_0$  (initial empty computation state)}
      \WHILE{$s \neq S$ \textbf{and} feasible}
      \STATE{ $s' \gets$ \textit{Choose next state from} $\mathcal{Y}(s)$ \color{orange}{using UCT}}
      \STATE{$x_{s'}$, feasible, {\color{orange}{reward}} $\gets o_{s, s'}(x_{s})$ }
      \STATE{\color{orange}\textit{Update UCT tree with} reward}
      \STATE{ $s \gets s'$}
    \ENDWHILE
    \IF{feasible}
        \STATE{append $x$ to $L$}
      \ENDIF
    \ENDWHILE
    \STATE  \textbf{Output:} List of valid samples $L$  
  \end{algorithmic}
\end{algorithm}



\subsection{Pruning the Sampling Tree using the Constraint Graph}

\begin{table}[t]
\centering
\caption{Number of transitions $s_i \to s_j$ in three scenarios of manipulation planning, before and after transition pruning.}
\label{tab:number-edges}
\begin{tabular}{@{}lccccc@{}}
\toprule
             & Variables & \begin{tabular}[c]{@{}c@{}}Transitions \\ Complete\end{tabular} & \begin{tabular}[c]{@{}c@{}}Transitions \\ after Pruning\end{tabular} & \begin{tabular}[c]{@{}c@{}} Pruning \\ Ratio [\%]\end{tabular}  \\ \midrule
Pick-Place  & 3      & 19                & 8    & 57.89 \\
Handover     & 7      & 2059                    & 163 &       92.09                               \\ 
Banana       &9       &    19171    & 534 & 97.21  \\
\bottomrule
\end{tabular}
\end{table}

Since the number of sequences and transitions grows exponentially with the dimension of the problem (see \cref{tab:number-edges}), it is not possible to naively apply MCTS on the previously defined MDP.
However, most real world problems have a sparse structure that can be leveraged to significantly reduce the set of available sequences and transitions.

We provide two examples for the transitions that can be pruned: 
Consider a pick and place problem represented with the variables $\{ q_1 , q_2 , t \}$, where $q_1, q_2$ are the robot configurations when picking and placing, and $t$ is the relative transformation between the end-effector and the object (also see the contraint graph in \cref{fig:pick_place}).
\begin{itemize}
  \item We prune $q_1 \to (q_1 , t)$ and $q_2 \to (q_2 , t)$.
  If $q_1$ or $q_2$ is sampled independently at random, the probability to grasp the object correctly is zero.
  \item We prune $t\to (q_1 , q_2)$ because $q_1$ and $q_2$ are independent if $t$ is fixed, and thus it does not make sense to generate $q_1$ and $q_2$ jointly.
\end{itemize}

Thus, we combine knowledge of the constraint graph structure  and the local information in each variable and constraint, i.e., the number of available degrees of freedom, and the number of equality constraints, to prune a transition $s_i \to s_j$  if it fulfills any of the two criteria: 

\paragraph{Zero probability of success}
In general, the probability of sampling variables from the ambient space such that equality constraints are fulfilled, is zero.
Hence, a transition can be pruned if the number of new, linearly independent, equality constraints exceeds the number of new degrees of freedom that are added in that transition.



\paragraph{Equivalence under conditional independence}
Variables in the constraint graph are conditionally independent if all paths that connect them with each other contain already assigned variables.
Given the partial assignment of a computational state $s_i$, we delete a transition that jointly samples variables that are conditionally independent with respect to the ones assigned in $s_{i}$.
In such a case, joint sampling means choosing an ordering of two conditionally independent processes, which is already represented in the computation graph

We show how the number of transitions is reduced in three example scenarios in robotics manipulation in \cref{tab:number-edges}.


\subsection{Problem Families and Tree Warmstart}
Obstacle and object configurations have an impact on the computational time and the success rate of the conditional sampling operations, potentially changing the optimal sampling sequence.
Thus, the best sampling order depends on both the constraint graph and the concrete problem instance.

MCTS provides a good framework to include the information gathered on solutions of similar problems as a warmstart.

We propose to warmstart the tree search by initializing $Q_i$ of each node with the average of the $Q$ values of the previous problems, and set $n_i$ to an \textit{equivalent count visit} $n_{\textrm{equiv}}$, that models how confident we are with the warmstart, following the approach presented in \cite{gelly2007combining}.



\section{Experimental Results}

\begin{figure}
\centering
\begin{subfigure}[t]{.16\textwidth}
    \centering
    \includegraphics[width=.9\linewidth]{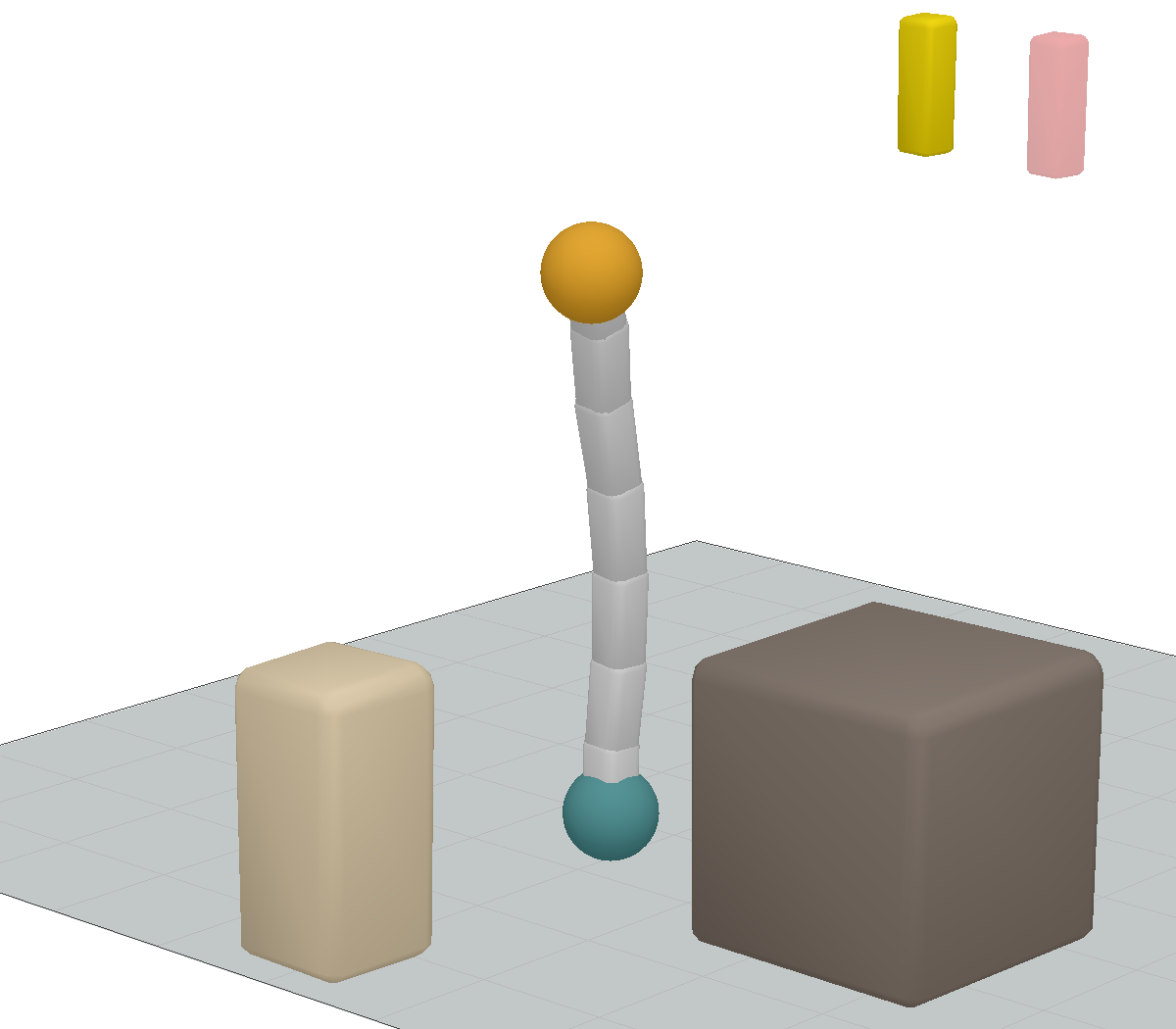} 
\end{subfigure}%
\begin{subfigure}[t]{.16\textwidth}
    \centering
    \includegraphics[width=.9\linewidth]{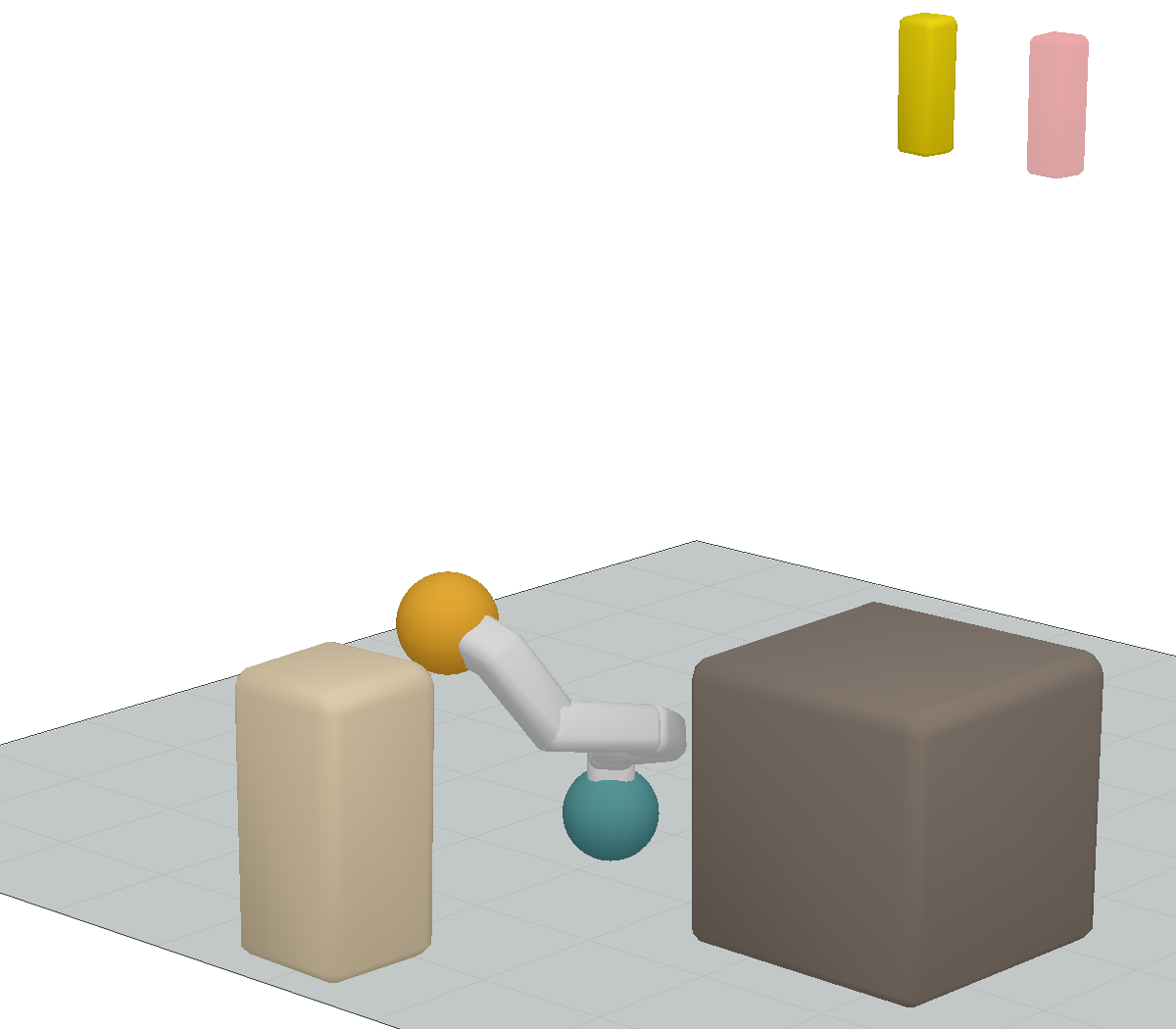} 
\end{subfigure}%
\begin{subfigure}[t]{.16\textwidth}
    \centering
    \includegraphics[width=.9\linewidth]{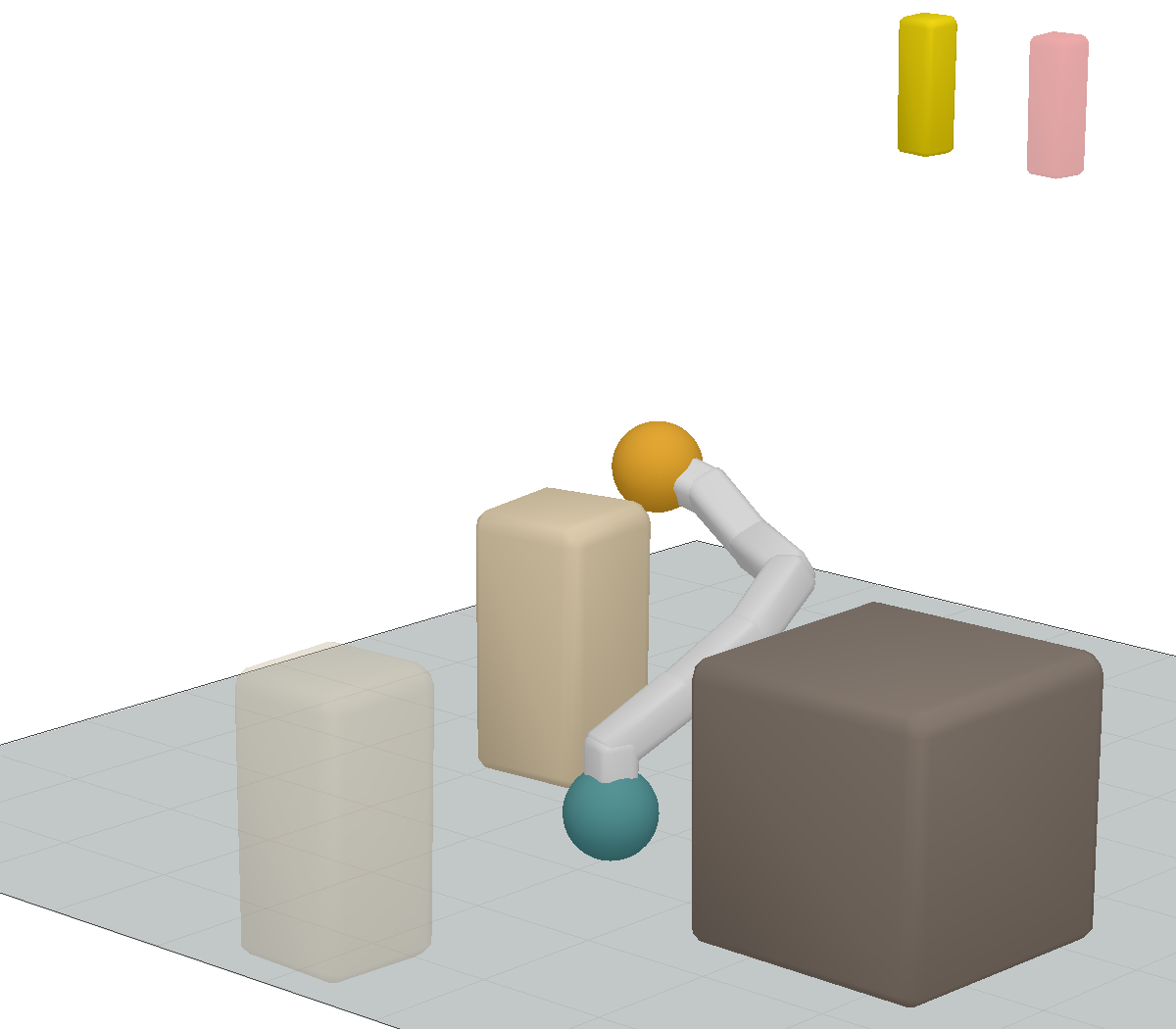} 
\end{subfigure}\\[1mm]

\begin{subfigure}[t]{.16\textwidth}
    \centering
    \includegraphics[width=.9\linewidth]{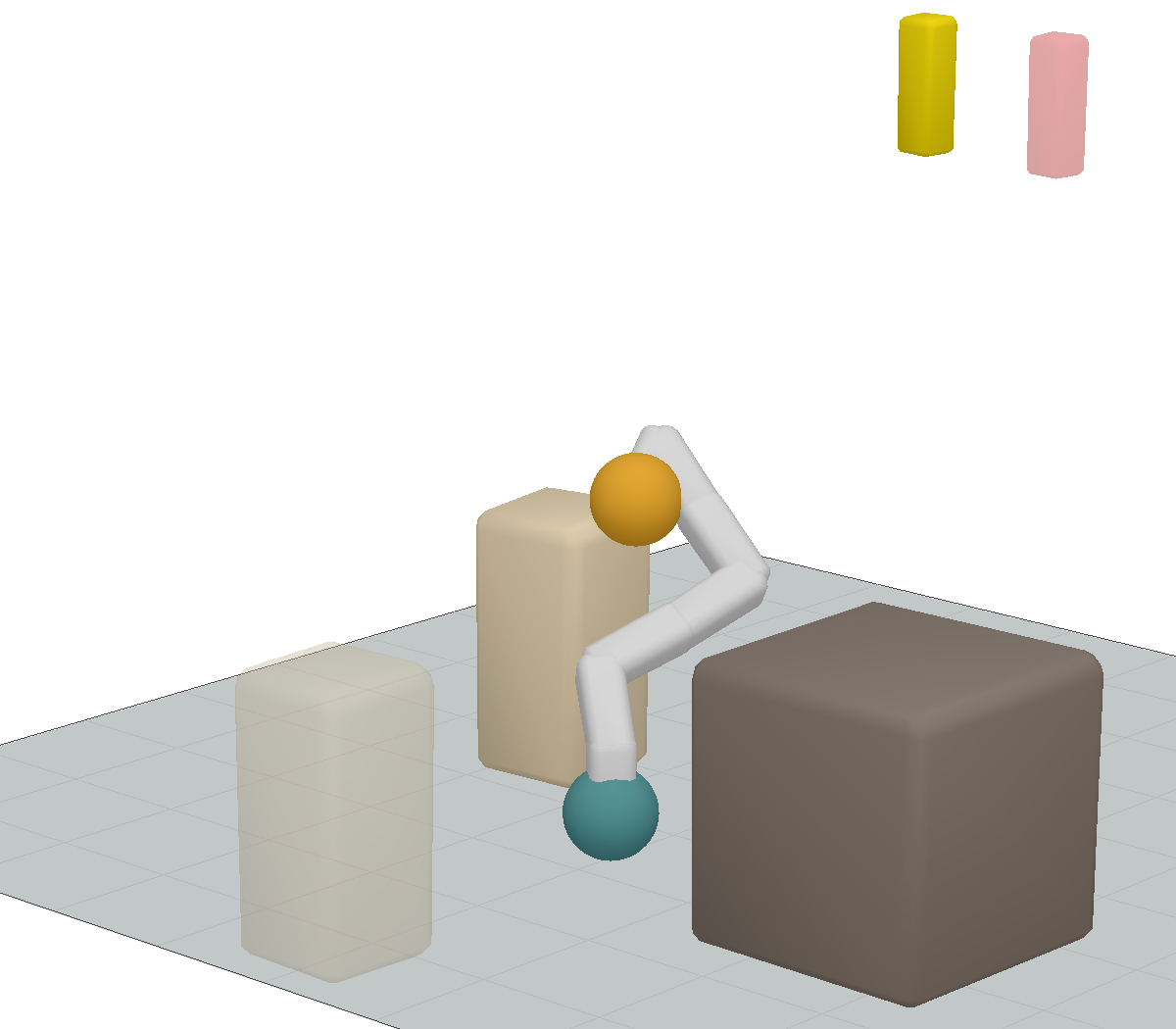} 
\end{subfigure}%
\begin{subfigure}[t]{.16\textwidth}
    \centering
    \includegraphics[width=.9\linewidth]{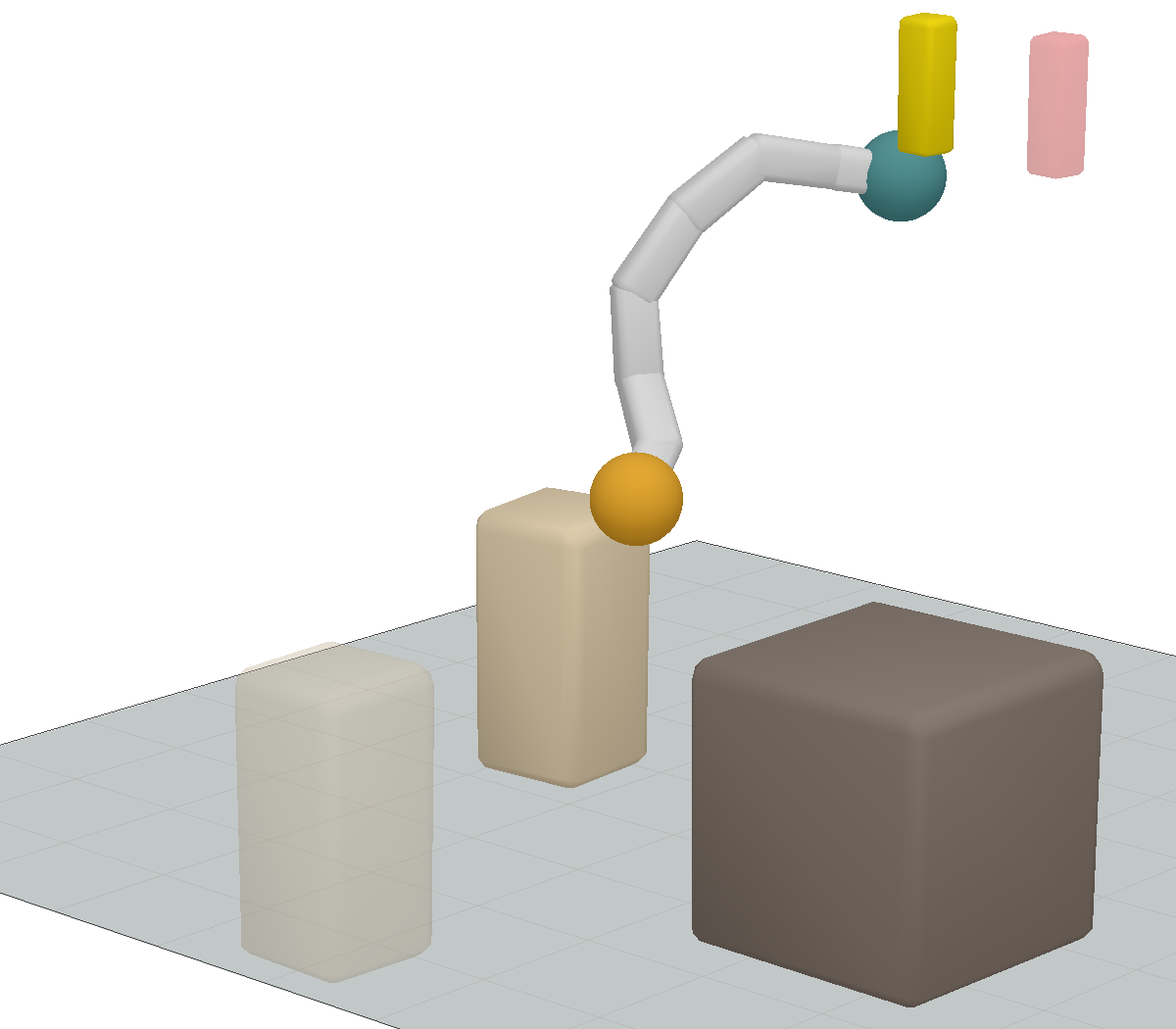} 
\end{subfigure}%
\begin{subfigure}[t]{.16\textwidth}
    \centering
    \includegraphics[width=.9\linewidth]{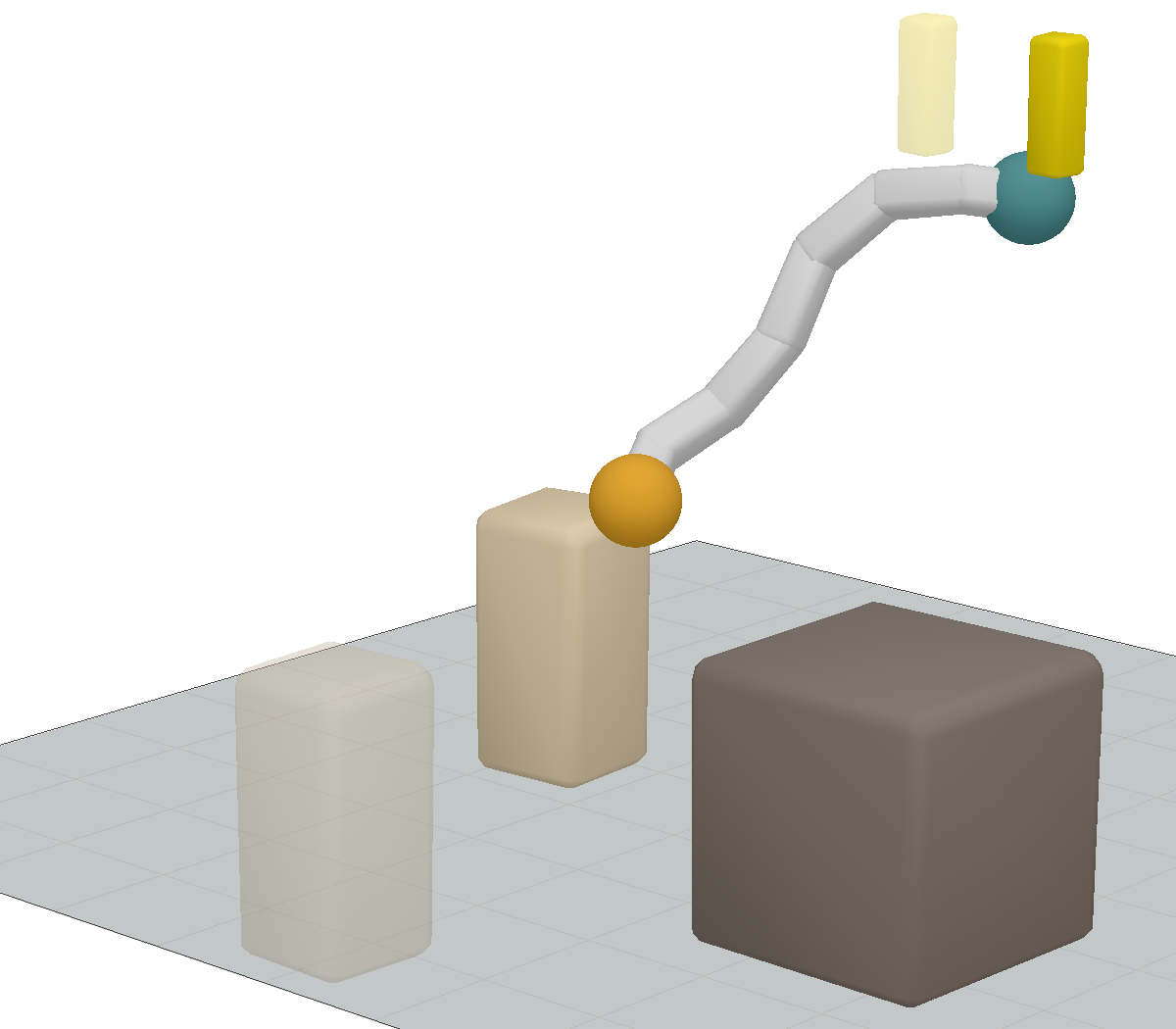} 
\end{subfigure}
    \caption{One possible sequence of mode switches in the \textit{banana} problem. The dark brown object is an obstacle.}
    \label{fig:keyframes-banana}
\end{figure}

\begin{table}[]
\setlength{\tabcolsep}{5pt}
  \centering
  \caption{Sampling sequences in \textit{Banana} and \textit{Handover} scenarios. Sampling order is left to right, each tuple $(.)$ denotes joint sampling of the subset. }
\label{tab:expert-sequences}
\begin{tabular}{@{}llr@{}}
\toprule
\multirow{2}{*}{\shortstack[l]{\textit{Expert 1} \\ Joint Opt.}} & Handover               & $(q_{a_1},
q_{a_2}, t_a , q_{b_1},  q_{b_2}, t_b , p )   $            \\ 
& Banana                   &
$( q_{a_1}, q_{a_2},   t_{a}, p ,q_{x},  t_{x},  q_{b_1}, q_{b_2}, t_{b})$
              \\ \midrule
\multirow{2}{*}{\shortstack[l]{\textit{Expert 2} \\ One-by-one}}& Handover  &  $p,t_a,t_b,q_{a_1},q_{a_2},q_{b_2},q_{b_1}$ \\
& Banana  &  $p,t_a,t_b,t_{x}, q_{a_1},q_{a_2},q_x,q_{b_1},q_{b_2}$ \\ \midrule
\multirow{3}{*}{\shortstack[l]{\textit{Expert 3}\\ Sequential\\ Mode-switches}}& Handover  & 
$( q_{a_1} , t_a), ( q_{a_2} , p, q_{b_2} , t_b ), (q_{b_1})$ \\
& Banana  & 
$(q_{a_1}, t_{a}),( q_{a_2}, p),( q_{x}, t_{x}),(q_{b_1}, t_{b}),q_{b_2}$ \\ \\ \midrule
\multirow{4}{*}{\shortstack[l]{\textit{Tree}: \\Best found \\Sequence \\(Examples)}}& Handover  &  $t_b, t_a, q_{a_1}, q_{b_1}, (q_{a_2},p),q_{b_2}$ \\
& & $(q_{b_1},t_b),(q_{b_2},p),(q_{a_2},t_a),q_{a_1}$ \\
& Banana  &  $(q_{a_1},t_a)$, $(q_{a_2},p),(q_{b_1},q_{b_2},t_b,t_x), q_x$ \\
& & $(p,q_{b_2},t_b,t_x)$,$q_x$,$(q_{a_1},t_a),$ $q_{a_2},$ $q_{b_1}$ \\ 
\bottomrule
\end{tabular}
\end{table}


\begin{figure*}
\centering
\begin{subfigure}[t]{.26\textwidth}
  \includegraphics[height=31mm]{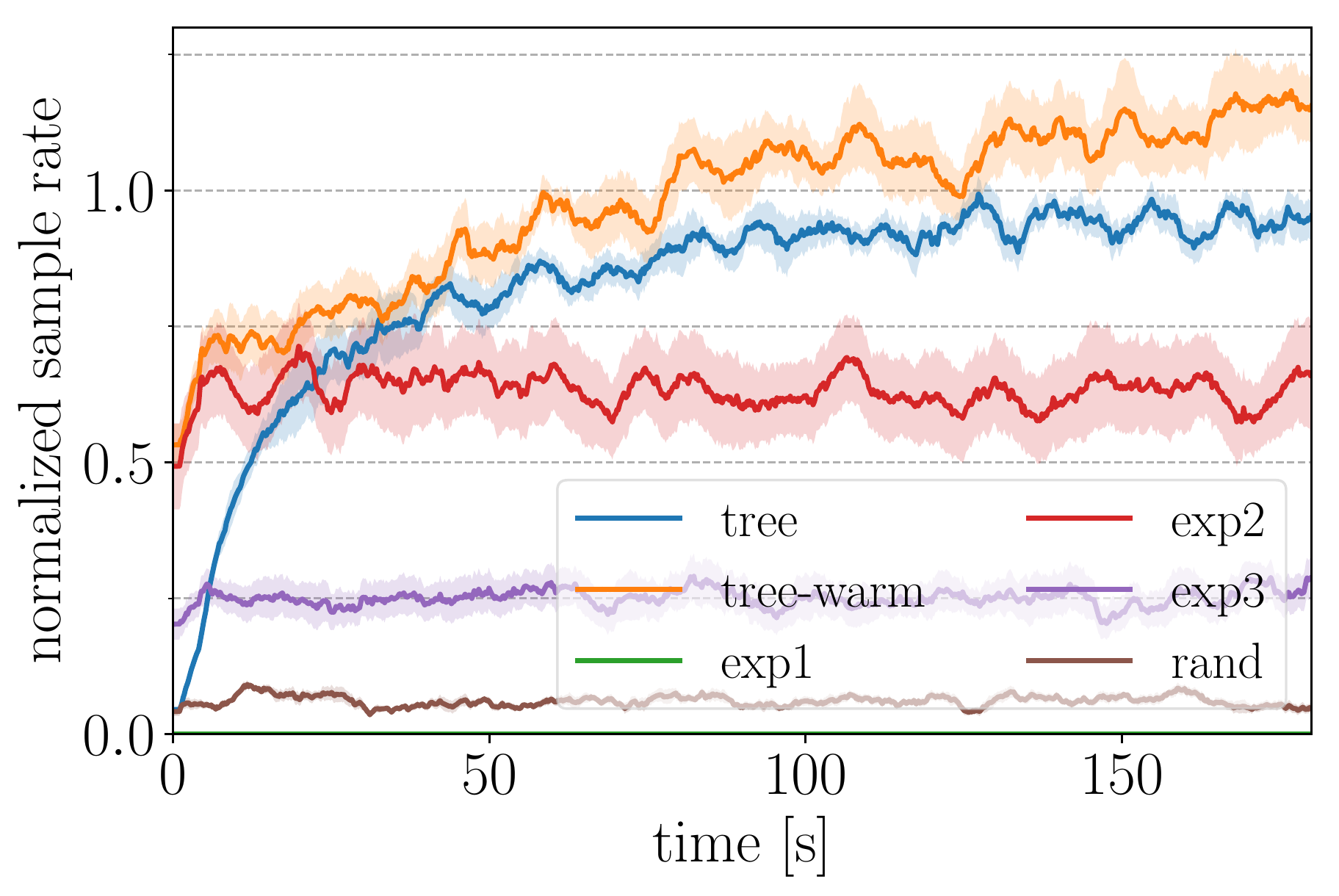}
  \caption{\textit{Handover}: Sample rate }
    \label{fig:handover_rate}
\end{subfigure}%
\begin{subfigure}[t]{.24\textwidth}
  \includegraphics[height=31mm]{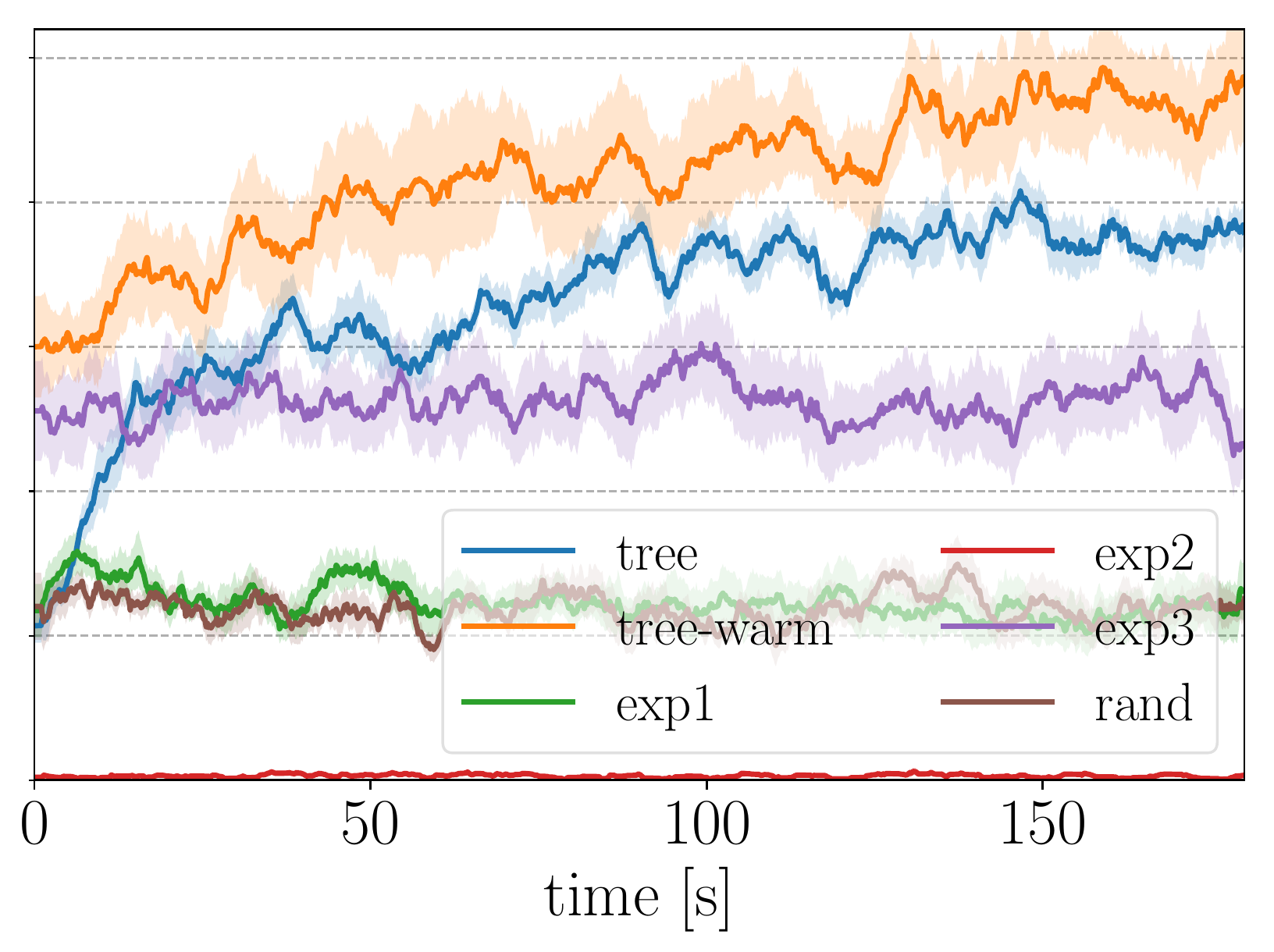}
  \caption{\textit{Banana}: Sample rate}
  \label{fig:banana_rate}
\end{subfigure}%
\begin{subfigure}[t]{.25\textwidth}
  \centering
  \includegraphics[height=31mm]{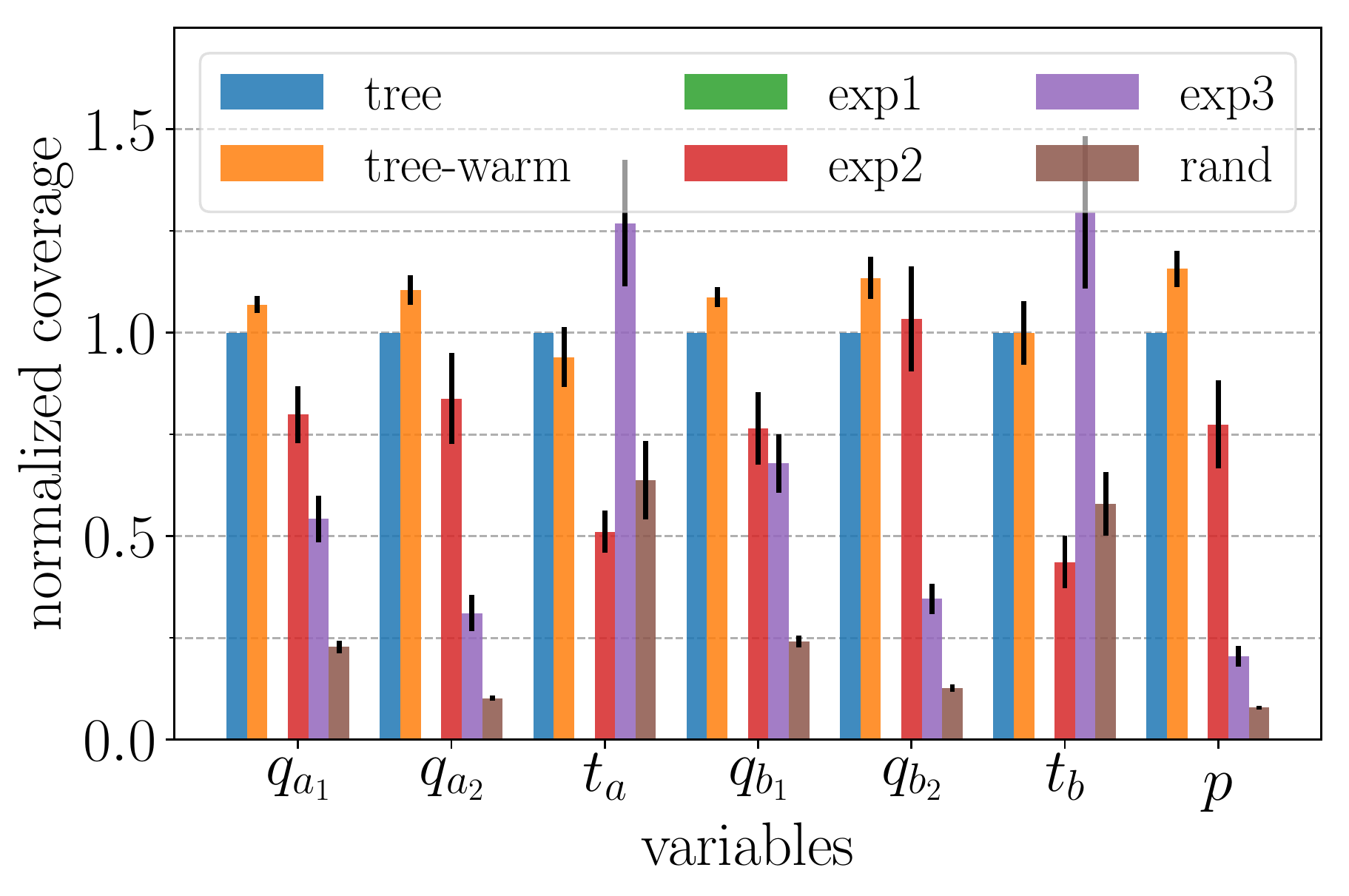}
  \caption{\textit{Handover}: Coverage }
  \label{fig:handover_coverage}
\end{subfigure}%
\begin{subfigure}[t]{.25\textwidth}
  \centering
  \includegraphics[height=31mm]{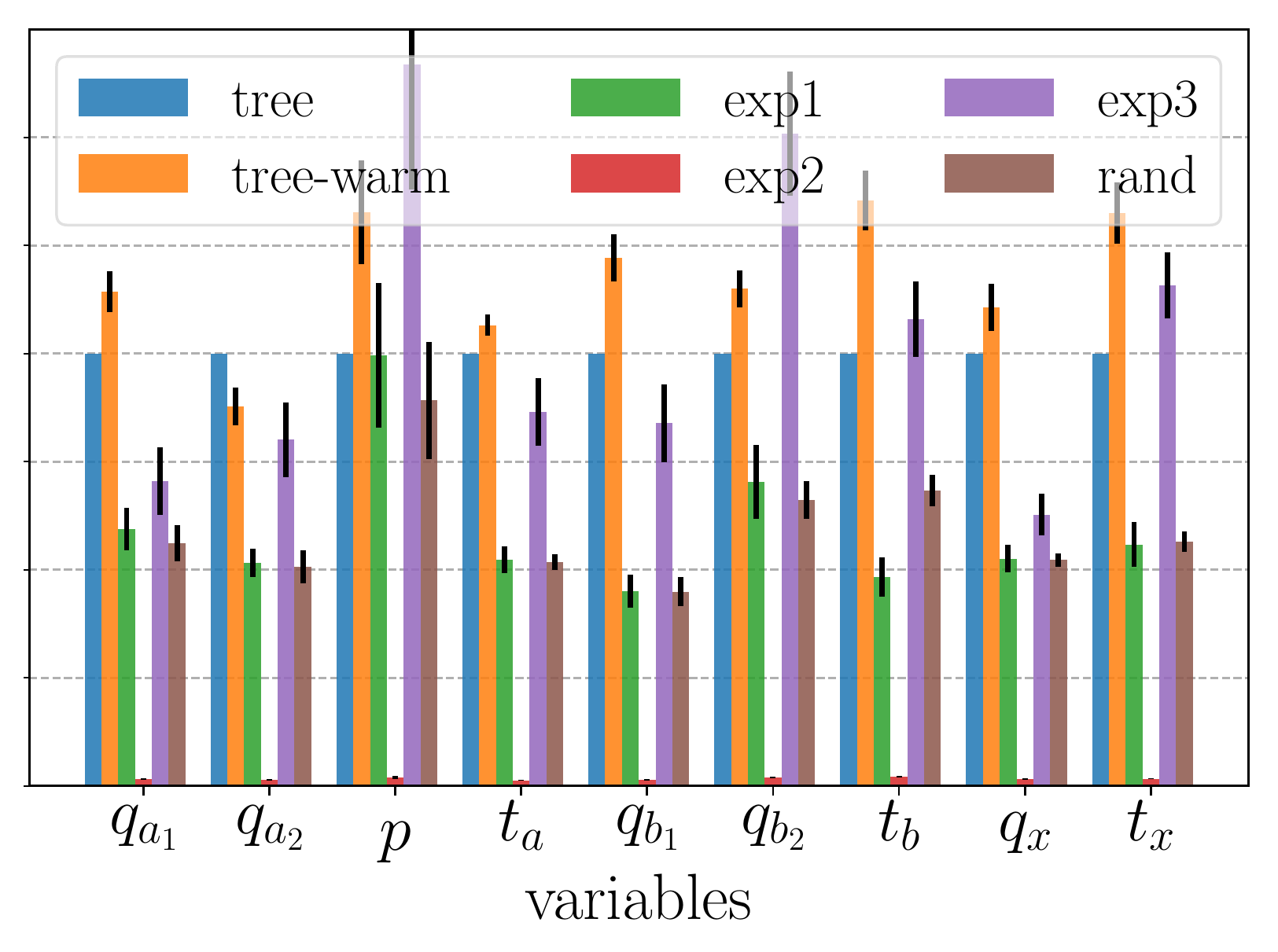}
  \caption{\textit{Banana}: Coverage}
    \label{fig:banana_coverage}
\end{subfigure}%
\caption{Sampling rates and approximated projected coverages (normalized by \textit{tree}), averaged over all problems instances.
}
\label{fig:exp}
\end{figure*}

\subsection{Scenarios}
We evaluate our approach on two complex sequential manipulation problems that pose different challenges in terms of nonlinearity and sparsity:

\begin{itemize}
    \item
    \textit{Handover}:
    Two robots (7D) have to collaborate to place a box in the goal configuration.
    Grasping is modeled with a two-finger gripper. 
    \Cref{fig:keyframes-handover} shows a possible solution and \cref{fig:factor-handover} shows the constraint graph.

    \item
    \textit{Banana:} The hanging `banana' has to be moved to a goal position.
    The robot (8D) can interact with the world using both sides of the kinematic chain as gripper.
    The box has to be moved, and used as a tool to stand on, to be able to reach the goal.
    Grasping is modeled as `grasp by touch'.
    \cref{fig:keyframes-banana} shows snapshots from a solution sequence and \cref{fig:banana-with-regrasp} shows the constraint graph.
\end{itemize}


For both problems, 8 different instances\footnote{For the problem instances, please check the accompanying video.} (i.e., varying obstacle and object configurations) are evaluated\footnote{Experiments are run using a Intel(R) Core(TM) i5-4200U 1.60GHz CPU.} multiple times, and if not stated otherwise, the average results are reported in the following analyses.
The different instances can have a big impact on the computation times, and thus the optimal sampling sequence.

\subsection{Computation operations}

One of the motivations of this work is that both the computation time  and the success rate of conditional sampling operations must be considered to design efficient algorithms. 
As an illustrative example, we report the average computational time and success rate of some sampling operations (which can return infeasible results) in the first instance of the \textit{handover} scenario:
Relative Transformation (0.68ms, 100\%), Inverse Kinematics (1.55ms, 87\%), Grasp of fixed object (5.8ms, 55\%), Pick  and Place (54ms, 46\%), Handover mode-switch (68ms, 44\%). 

In practice, our algorithm finds the optimal balance between \textit{(i)} choosing simple operations which are fast, and have a high success rate, but can potentially induce infeasibility in future assignments and \textit{(ii)} optimizing variables jointly,
which considers joint feasibility but is slower and often has a lower success rate.


The conditional sampling operations are implemented  by randomizing the initial guess of a nonlinear solver without cost term, and solved using KOMO \cite{14-toussaint-KOMO}.
The initial guess covers the whole ambient space of the current partial assignment such that all possible partial solutions have a non-zero probability 
(see \cref{sec:framework}).




\subsection{Number of Samples and Approximate Coverage} 


We compare the number of samples generated by our algorithms (\textit{tree} and \textit{tree-warm}), against \textit{expert}-sequences (see \cref{tab:expert-sequences} for the specific sequences) and \textit{Rand}.
Expert-sequences are fixed sampling orders that represent different user-defined strategies.
\textit{Rand} chooses computational operations at random, without learning.

\paragraph{Number of samples}
We plot the evolution of the sampling rate (including MCTS overhead for our algorithms) in our two scenarios in \cref{fig:handover_rate,fig:banana_rate}. 
Asymptotically, \textit{tree} outperforms all experts, which have a disparate performance across our two scenarios. 
Due to the exploration of possible sampling sequences, in the beginning the sample rate of \textit{tree} is lower compared to some of the \textit{expert}-sequences. 
On the other hand, $\textit{tree-warm}$ alleviates this issue due to the warmstart, and improves its sample rate as time advances. 
We also observe that the \textit{expert}-sequences are more sensitive to the different problem instances.

The final sample rates achieved by \textit{tree-warm} on the different instances vary between 7-15 samples/s in \textit{handover} and 5-10 samples/s in \textit{banana}.
We show examples for some of the discovered sequences by \textit{tree} in \cref{tab:expert-sequences}.


\paragraph{Approximate Coverage}
We evaluate the coverage of the solution manifold achieved by our algorithms \textit{tree} and \textit{tree-warm}, which should achieve good coverage (i.e., a diverse set of samples) by design.
Since the coverage of a nonlinear manifold embedded in a high-dimensional space can not be evaluated reliably with a relatively low number of samples, we evaluate the `projected coverage' as a proxy-measure:
We project the full samples to each variable of the sequence
and compare the coverage in each of these subspaces.
We discretize these subspaces and count the number occupied cells, i.e., cells that are occupied by multiple samples are only counted once.
Although an approximation, it provides useful and interpretable information about which parts of the solution-manifold are covered. 

The results shown in \cref{fig:handover_coverage,fig:banana_coverage} are normalized by the number of occupied cells in \textit{tree}.
While some experts achieve better coverage on a subset of variables, coverage of \textit{tree} and \textit{tree-warm} mostly outperform the others. 
This confirms our hypothesis that, with correct design and implementation of the sampling operations, maximizing the number of samples is a good heuristic to maximize the coverage.

\section{Conclusion}

We proposed a framework to reason about optimal decompositions of factored nonlinear programs and demonstrated it on feasibility problems arising in robotics manipulation planning.
Our algorithm chooses the computational decisions, i.e., which subset of variables to conditionally sample next, to maximize the number of generated samples in a fixed computational time.
We use the method to efficiently generate a diverse set of samples for mode-switches in robotic manipulation, which is essential for guiding trajectory optimization, or to set goal and start configurations for sampling-based motion planning.
A limitation is that the algorithm must explore inefficient computations  and needs some time to converge to the best sampling order, which could prevent usage when the computational budget is very limited.
As demonstrated, this is alleviated by warmstarting the algorithm using information from previous runs.

Our framework would naturally allow to also include cost factors in the NLP. We want to briefly comment on why we chose to neglect cost terms: Our approach is tailored to provide a diverse set of feasible samples that can be used to seed higher-level optimization or path finding problems, in particular, to escape potential infeasible local optima.
The diversity and uniform coverage of samples is an essential ingredient for the completeness of such overall solvers. Including a cost term in the NLP would be straight-forward but compromise the diversity of generated samples and make the overall approach again prone to local optima.

\bibliographystyle{IEEEtran}
\typeout{} 
\bibliography{IEEEabrv,mybibfile}


\end{document}